\documentclass[preprint,12pt]{elsarticle}

\usepackage{amssymb}

\usepackage{hyperref}
\usepackage{multirow}
\usepackage{booktabs}
\usepackage{amsmath}
\usepackage{colortbl}
\usepackage{tabularx}
\usepackage{rotating}
\usepackage{graphicx,subcaption,ragged2e}
\usepackage{adjustbox}
\usepackage{setspace}

\journal{Pattern Recognition}

\begin{document}

\begin{frontmatter}

\title{Reinforcement Learning-based Mixture of Vision Transformers for Video Violence Recognition}

\author[inst1]{Hamid Mohammadi}
\ead{hamid.mohammadi@aut.ac.ir}

\affiliation[inst1]{organization={Computer Engineering Department, Amirkabir University of Technology},
            city={Tehran},
            state={Tehran},
            country={Iran}}

\author[inst1]{Ehsan Nazerfard \corref{cor1}}
\ead{nazerfard@aut.ac.ir}
\author[inst2]{Tahereh Firoozi}
\ead{tahereh@ualberta.ca}

\affiliation[inst2]{organization={Measurement, Evaluation, and Data Science (MEDS), University of Alberta},
            city={Edmonton},
            state={Alberta},
            country={Canada}}

\cortext[cor1]{Corresponding author}

\begin{abstract}
Video violence recognition based on deep learning concerns accurate yet scalable human violence recognition. Currently, most state-of-the-art video violence recognition studies use CNN-based models to represent and categorize videos. However, recent studies suggest that pre-trained transformers are more accurate than CNN-based models on various video analysis benchmarks. Yet these models are not thoroughly evaluated for video violence recognition. This paper introduces a novel transformer-based Mixture of Experts (MoE) video violence recognition system. Through an intelligent combination of large vision transformers and efficient transformer architectures, the proposed system not only takes advantage of the vision transformer architecture but also reduces the cost of utilizing large vision transformers. The proposed architecture maximizes violence recognition system accuracy while actively reducing computational costs through a reinforcement learning-based router. The empirical results show the proposed MoE architecture's superiority over CNN-based models by achieving 92.4\% accuracy on the RWF dataset.
\end{abstract}

\begin{highlights}
    \item Vision transformers achieve state-of-the-art accuracy in video violence recognition
    \item Large state-of-the-art vision transformers are computationally costly
    \item Various transformers can be combined for cost-efficient video violence detection
    \item A mixture of transformers system can be built using a reinforcement learning router
\end{highlights}

\begin{keyword}
Video Violence Recognition \sep Vision Transformers \sep Mixture of Experts \sep Deep Reinforcement Learning
\end{keyword}
\end{frontmatter}

\section{Introduction}
\label{sec:introduction}

Video violence recognition as a sub-task of video action recognition requires accurate and detailed human motion analysis. To accurately recognize violence in videos, contextual motion semantics are as significant as its physical properties such as direction and speed \cite{mumtaz2022overview}. As a result, combining contextual information from the current environment and previous frames is necessary to understand the meaning of the present motion \cite{kong2022human}. To capture spatial-temporal information from a 3D video, models that can capture the spatiotemporal relationship between different regions of a video are used \cite{hao2022spatiotemporal}. There are, however, several challenges associated with this problem, including spatiotemporal pattern complexity, video length, difficulty gathering an extensive and comprehensive video violence dataset, and redundancy of information in the video dataset \cite{mohammadi2023video, zheng2020dynamic}.

Convolutional models, the dominant models in computer vision, have also been heavily used to recognize violence in video \cite{mohammadi2023video, islam2021efficient, su2020human, cheng2021rwf}. An early example of such methods is to combine 2D convolutional backbones with recurrent feature aggregators \cite{ge2019attention, zhang2020human}. A significant improvement in video classification models was achieved by using 3D convolutional layers \cite{huang2023review}. Additionally, 3D convolutional backbones eliminated the need for recurrent aggregation of features, which contributed to the stability of the training video model \cite{portsev2021comparative}. Furthermore, auxiliary features were introduced to extract more useful features from video clips, including optical flow \cite{wang2019hallucinating}, skeleton estimations \cite{hachiuma2023unified, duan2022revisiting}, and deep features \cite{hung2021violent, honarjoo2021violence}. Although CNN-based video violence recognition models are becoming more accurate, their computational complexity and learning capacity challenge their further adaptation \cite{mumtaz2022overview}.

Vision transformers are a novel paradigm introduced as an alternative method of capturing spatiotemporal information in 3D video data. Vision transformers and CNNs differ fundamentally in using self-attention mechanisms instead of convolutional layers for pattern recognition in videos \cite{pan2022integration}. Unlike convolutional filters, self-attention can recognize long-distance relations between image patches without relying heavily on multilayer architectures. As a result, vision transformers are able to capture global features more efficiently \cite{ramachandran2019stand, vaswani2021scaling}. In addition, the self-attention mechanism implementation is more hardware friendly, which means that current deep-learning accelerators can handle these computations more efficiently \cite{mumtaz2022overview}.

Self-attention mechanisms lack the inductive bias found in convolution filters \cite{tuli2021convolutional}. Although the inductive bias of convolutional filters is important when learning viable representations from small amounts of visual data, it is limiting when large amounts of data are available \cite{dosovitskiy2020image}. As a result, while CNN-based models have greater performance than vision transformers when applied to low-data scenarios, vision transformers have superior performance when applied to large datasets \cite{dosovitskiy2020image}. Accordingly, vision transformers can exploit large datasets more effectively than CNN-based models \cite{khan2022transformers}. Self-supervised training illustrates the significance of this difference. While self-supervised training is possible in CNN-based models \cite{ge2021revitalizing}, vision transformers can significantly outperform CNN-based methods when pre-trained with large amounts of unsupervised visual data \cite{tong2022videomae}.

\section{Literature Review}
\label{sec:lit:review}

Currently, violence recognition and action recognition methods can be divided into two groups: CNN-based and Transformer-based. Convolutional neural networks, due to their inherent inductive bias, can learn rich visual features from images/videos even with limited visual data \cite{dosovitskiy2020image}. Additionally, CNNs benefit from transfer learning via supervised pre-training to downstream tasks \cite{ge2021revitalizing}. However, with the steady increase in visual data, vision transformers showed their potential because they have a lower inductive bias and more parameters. In addition, the transformer's scalability allows it to use massive bodies of unannotated visual data for model pre-training via self-supervised training. Through unsupervised attainment of rich visual presentation, vision transformers are able to increase their performance ceiling for downstream visual tasks \cite{tong2022videomae}.

The convolutional network processes video as either a sequence of 2D data (images) or 3D data. For the sequence of 2D data assumption, CNN layers are used to extract semantic features from each image. The extracted features are then aggregated using aggregators like temporal pools \cite{wang2021spatial} or recurrent networks \cite{ge2019attention, zhang2020human}. While these approaches are suitable for smaller datasets, their inductive bias makes them unsuitable for scaling with more data \cite{dosovitskiy2020image}. By contrast, 3D convolutions are based on 3D filters that identify spatiotemporal features within videos, delivering temporally aware features \cite{suzuki2018learning}. In the presence of a large enough dataset, 3D convolutional networks perform better than their 2D counterparts \cite{jiaxin2021review}. Thus, despite their computational requirements, many modern video analysis methods rely heavily on 3D convolutional neural networks \cite{qiu2019learning, feichtenhofer2019slowfast}.

Additional useful information is added to CNN-based networks' features by explicitly extracting application-specific information for the action recognition task. Optical flow \cite{sevilla2019integration, wang2019hallucinating}, skeleton estimation \cite{hachiuma2023unified, duan2022revisiting}, and object-aware deep features \cite{hung2021violent, honarjoo2021violence} are the most common examples of such information. While this information is implicit in raw video data, its explicit inclusion facilitates action recognition by increasing the density of input information. In spite of the advantages of such auxiliary information, these methods are unfavorable because of the annotated data and computational requirements \cite{mohammadi2023video}.

CNN-based methods can also benefit from soft and hard attention to improve input data purity. Soft attention weights data parts according to their importance, which reduces the noise-to-data ratio \cite{niu2021review}. Meanwhile, hard attention can be viewed as a cut-off version of soft attention that removes the less significant parts of the input data. As a result, computational resources are saved since less significant parts aren't processed \cite{yang2020overview, sun2020introductory, mohammadi2023video}. Furthermore, both soft and hard attention can be used to localize the region of interest within the visual data \cite{mohammadi2023video}.

The prominent state-of-the-art CNN-based video action recognition models, namely SlowFast \cite{feichtenhofer2019slowfast} and LGD-3D Two-stream \cite{qiu2019learning} models are multi-pathway networks aimed at optimal utilization of different axes of information in a video clip. Each pathway focuses primarily on extracting local, global, or temporal information. The SlowFast network consists of two pathways, one extracting fine-grain spatial features and one focused on temporal feature extraction. The spatial pathway operates on fewer high-resolution frames to capture semantic features. On the other hand, the temporal pathway extracts temporal information from a larger number of frames. The LGD (Local Global Diffusion) network employs different pathways to extract and combine local and global features. The local pathway uses 3D convolutional filters to capture high-resolution features. In contrast, the global pathway operates on a downscaled version of input frames obtained via Global Average Pooling (GAP) to capture the relation between distant patches of pixels. Using specialized spatial or temporal pathways, both approaches attempt to compensate for convolutional filters' lack of understanding of long distant relationships.

This study focuses on the Real World Fights (RWF) dataset since it is the largest and most diverse dataset available for video violence recognition \cite{cheng2021rwf}. Additionally, \cite{cheng2021rwf} introduced a CNN-based model called Flow Gated Network (FGN), utilizing the model to recognize violence in videos. FGN uses a two-stream architecture to process RBG frames and optical flow frames. To increase the information density of their models' input, they used a motion-based hard attention method, similar to \cite{wang2016beyond}. The features extracted by the RGB and optical flow backbones are fused together with a learned weighted sum. This forms a global feature that is classified via an MLP. This model has achieved a maximum accuracy of 87.25\% on the RWF dataset \cite{cheng2021rwf}.

Violence detection can be made simpler and more generalizable by using explicit features such as skeleton estimates. \cite{su2020human} has proposed the Skeleton Points Interaction Learning (SPIL) model. This paper proposes recognizing the violent or non-violent nature of action based on the interaction between 3D skeleton points. It identifies connections between the 3D skeleton points derived from the input video frames. Such methods produce noise-free information regarding the performed task by explicitly extracting critical information about human movements. Skeletons extraction, however, requires additional computations and training and might not be highly accurate in every situation. The SPIL model achieved 89.3\% accuracy on the RWF dataset \cite{su2020human}.

\cite{islam2021efficient} focuses on the video violence detection task performance efficiency. The paper uses a two-stream architecture using mobile-net backbones to process two sets of information. These backbones process RBG-difference frames and background-suppressed RBG frames. The background-suppressed RGB frames are obtained using the ViBe background estimation method \cite{barnich2009vibe}. The suppression of background data helps to maintain the purity of input data. In addition, a simple RGB-difference method can extract temporal information between adjacent frames by comparing the corresponding RBG value of pixels. The paper uses separable convolutions to reduce computation costs during convolutional backbone feature extraction. LSTM networks are used to aggregate video features between frames. The proposed model, called SepConvLSTM, achieves an accuracy of 89.75\% on the RWF dataset \cite{islam2021efficient}.

Increasing data purity is a common theme in state-of-the-art violence recognition research. \cite{mohammadi2023video} have solely focused on improving accuracy and performance via a semi-supervised method based on deep reinforcement learning. The study dismisses the use of auxiliary features for accuracy improvement because of their negative effect on model performance. However, removing redundant data through hard attention allowed the model to focus on the region of interest at a higher resolution. As a result, the model can achieve higher accuracy by processing high-resolution footage without increasing model size, keeping it agile. Moreover, using a learned region of interest localization strategy instead of heuristic-based motion filtering or background suppression increases the precision of region of interest localization and avoids false positives. The model uses a pre-trained I3D \cite{carreira2017quo} backbone trained on the Kinetics dataset \cite{kay2017kinetics}. Pre-training significantly reduces training time spent on the downstream task. Their model, Semi-Supervised Hard Attention (SSHA), achieved the current state-of-the-art accuracy of 90.4\% on the RWF dataset \cite{mohammadi2023video}.

\begin{table}[!htbp]
\centering
\caption{Accuracy of state-of-the-art CNN-based models on the RWF dataset.}
\label{tab:compare:rwf}
\noindent
\begin{tabular}{lc}
\toprule
\textbf{Model} & \textbf{Accuracy (\%)} \\
\midrule

Flow Gated Network \cite{cheng2021rwf} & 87.2 \\
SPIL \cite{su2020human} & 89.3 \\
SepConvLSTM \cite{islam2021efficient} & 89.7 \\
SSHA \cite{mohammadi2023video} & 90.4 \\

\bottomrule

\end{tabular}
\end{table}

Despite the success of the CNN-based solutions family, drawbacks also exist. Although parameter sharing in CNN-based models makes them efficient regarding memory usage, convolutional operations are computationally heavy \cite{alzubaidi2021review}. This problem is more tangible when applying these models to high-resolution visual data \cite{alzubaidi2021review, sun2022computation}. Additionally, state-of-the-art solutions sometimes use additional features to improve CNN-based models' spatiotemporal understanding. These features are computationally heavy to extract and require feature-specific backbones to process such auxiliary features. For example, hardware-accelerated algorithms are necessary to compute the optical flow feature rapidly and efficiently. On top of that, a computationally heavy multi-stream network is used to utilize this feature besides the original RGB frames. The use of auxiliary features such as skeleton estimations also requires additional supervised training data, which increases the cost of using such techniques.

The inductive bias of CNN-based models is both an advantage and a disadvantage. CNN-based models can show impressive learning and generalization abilities when trained on a limited set of visual data \cite{dosovitskiy2020image}. However, their scaling is quickly saturated when trained on larger vision datasets \cite{dosovitskiy2020image}. The problem limits the use of CNN-based models for applications that require higher levels of accuracy, as well as the use of unsupervised and self-supervised methods. As CNN-based models are limited when it comes to using a large body of visual data due to their inductive bias, there is limited use in using unsupervised pre-training to create rich internal visual representations in these models \cite{chen2021vision, zhai2022scaling}.

The vision transformers \cite{dosovitskiy2020image} follow in the footsteps of natural language processing \cite{vaswani2017attention} in their approaches to visual understanding. Instead of applying locally focused filters to image pixels (i.e., convolutional filters), transformers use self-attention to capture local and global relations between image patches \cite{ramachandran2019stand, vaswani2021scaling}. Using self-attention to process visual patches and the inter-patch relations reduces inductive bias in this architecture family. Transformers perform poorly when there are no large datasets, but their learning ability increases when more data is available \cite{dosovitskiy2020image}. Additionally, the modular treatment of visual patches in transformers creates the opportunity of developing flexible methods for transformers un- and self-supervised training and performance-accuracy trade-off management \cite{li2022exploring}. Furthermore, transformers' built-in attention mechanism adds the advantage of explicit input patch prioritization and increased explainability to these models.

Video vision transformers are a natural extension of the image vision transformers concept with minimal architectural modification. Whether a video vision transformer uses 2D or 3D visual data patches, the patch encoding can be processed by a vision transformer by adding suitable spatial, temporal, or spatiotemporal positional encoding \cite{bertasius2021space}. This flexibility allows applying pre-trained image transformers to video analysis problems \cite{arnab2021vivit, bertasius2021space}. Additionally, it reduces video transformers' dependency on regular time intervals between consecutive input frames, a drawback of 3D convolutions \cite{savran2021novel}.

"Is time-space attention all you need for video understanding?". Bertasius et al. \cite{bertasius2021space} answer this question by introducing an alternative approach to spatiotemporal data understanding via self-attention. TimeSFormer \cite{bertasius2021space} uses self-attention to understand the connection between image patches distributed across the space and time axis. The paper has three main takeaways: (i) Due to the large number of parameters in a vision transformer, a considerable amount of pre-training is required for vision transformers to perform equally or better than their CNN-based counterparts. This is both negative and positive. The negative aspect is that transformer-based video analysis models require more data to perform better than CNN-based models. However, transformers can benefit from more training data without saturation. (ii) TimeSFormer paper shows that this model obtains the highest action recognition accuracy using a separate spatial and temporal attention mechanism. The alternative method uses joint spatiotemporal attention to connect image patches across space and time axes. However, due to the exponential complexity of joint spatiotemporal attention, it is prone to overfitting given the limited amount of video data available at this point \cite{bertasius2021space}. (iii) And finally, although transformers are larger in terms of the number of parameters, vision transformers are computationally more efficient \cite{pan2022edgevits}.

Vision transformers' lower inductive bias can also be applied to video analysis. VideoMAE \cite{tong2022videomae} proposed a technique similar to Masked Language Modeling (MLM) \cite{devlin2018bert} for Self-Supervised Video Pre-training (SSVP). This paper draws inspiration from \cite{he2022masked} that uses a similar technique for Self-Supervised Image Pre-training (SSIP). The SSVP technique trains a video encoder-decoder by masking a large (90\% to 95\%) portion of input video tubes (spatiotemporal regions). The model learns useful video representations by predicting masked portions of videos. VideoMAE demonstrates the impressive ability of vision transformers to achieve favorable results on small video datasets only using SSVP. As presented in this paper, the huge learning potential of vision transformers makes it possible to combine SSVP techniques and supervised pre-training to achieve new state-of-the-art results on common video action and violence recognition datasets.

Table \ref{tab:benchmark:kinetics} compares state-of-the-art video action recognition models from self-attention-based and convolution-based origins. VideoMAE \cite{tong2022videomae} with different sizes (ViT-H, -L, -B, and -S) and TimeSFormer \cite{bertasius2021space} with different sizes (L, HR, and base) are representatives of the self-attention-based branch of video action recognition models. Moreover, LGD-3D \cite{qiu2019learning}, SlowFast \cite{feichtenhofer2019slowfast}, and I3D-NL \cite{wang2018non} are representatives of convolutional-based action recognition models.

To summarize the information in Table \ref{tab:benchmark:kinetics}, the VideoMA-H model achieves the highest top 1 accuracy on the kinetics 400 dataset (86.6\%) compared to the other self-attention-based and convolution-based models. In contrast, the I3D model achieved the least accuracy on the same dataset (71.1\%). However, when considering the computation and memory footprint of the models, the VideoMAE-H model also has the highest number of parameters (633M) and the largest required GFLOPs for inference (17.8 TFLOPs), while I3D has the lowest number of parameters (12.2M) and TimeSFormer-B has the smallest required GFLOPs (0.6 TFLOPs). The number of sampled frames from a video clip varies greatly among the models, with TimeSFormer-L having the most input frames (96) and TimeSFormer-H, TimeSFormer-B, and LGD-3D Two-stream having the lowest number of input frames (8). The input size of all the mentioned models is similar (224x224) except for TimeSFormer-HR (448x448) and the SlowFast model (256x256).

Two main takeaways from Table \ref{tab:benchmark:kinetics} are: (i) Transformers can utilize their higher learning saturation cap to scale up using larger unsupervised and supervised training sets. The VideoMAE model achieves significantly higher accuracy on the Kinetics 400 dataset utilizing self-supervised pre-training made possible by its masked auto-encoder pre-training. Additionally, transformers could be computationally lighter even though they have a larger number of parameters. For example, the TimeSFormer-B model can achieve roughly the same accuracy (78\%) as the LGD-3D Two-stream (81.2\%) and SlowFast (79.8\%) models with lower TFLOPs of computation (TimeSFormer-B = 0.6, LGD-3D Two-stream = N/A, and SlowFast = 7.0) despite having a larger number of parameters (TimeSFormer-B = 121.2, LGD-3D Two-stream = 34.9, and SlowFast = 34.3). Transformers can use hardware resources more efficiently as the image patching and self-attention algorithms are less computationally complex than the convolutional operation, hence requiring fewer TFLOPs to process an identical input \cite{pan2022edgevits}.

\begin{sidewaystable}[!hp]
\centering
\caption{State-of-the-art video classification models' properties and accuracy on Kinetics-400 \cite{kay2017kinetics} dataset. Vision transformers show superior performance on the Kinetics-400 dataset.}
\label{tab:benchmark:kinetics}
\begin{tabularx}{\textwidth}{XXXXXXXX}
    \toprule
\textbf{Class} & \textbf{Model} & \textbf{\# parameters (M)} & \textbf{TFLOPs} & \textbf{\# input frames} & \textbf{Frame size} & \textbf{Top 1 accuracy (\%)} & \textbf{Top 5 accuracy (\%)}  \\
    \midrule
\multirow{3}{2cm}{\textbf{Self-attention-based}}
& VideoMAE (ViT-H) & 633 & 17.8 & 16 & 224x224 & 86.6 & 97.1 \\
& VideoMAE (ViT-L) & 305 & 8.9 & 16 & 224x224 & 85.2 & 96.8 \\
& VideoMAE (ViT-B) & 87 & 2.7 & 16 & 224x224 & 81.5 & 95.1 \\
& VideoMAE (ViT-S) & 22 & 0.9 & 16 & 224x224 & 79.0 & 93.8 \\
& Timesformer (L) & 121.3 & 7.1 & 96 & 224x224 & 80.7 & 94.7 \\
& Timesformer (HR) & 121.7 & 5.1 & 8 & 448x448 & 79.7 & 94.4\\
& Timesformer (B) & 121.2 & 0.6 & 8 & 224x224 & 78 & 93.7 \\
\cmidrule(l){2-8}
\multirow{3}{2cm}{\textbf{CNN-based}}
& LGD-3D Two-stream & 34.9 & N/A & 8 & 224x224 & 81.2 & 95.2 \\
& SlowFast R101-NL & 34.3 & 7.0 & 16 and 8 & 256x256 & 79.8 & 93.9 \\
& I3D-NL & 12.2 & 10.8 & 64 & 224x224 & 77.3 & 93.3 \\
    \bottomrule
\end{tabularx}
\end{sidewaystable}

This study evaluates state-of-the-art video transformers on video violence recognition tasks to assess their effectiveness in this subject. Additionally, an efficient method to combine the accuracy of large vision transformers with the low computational cost of smaller vision transformers is presented. The proposed method uses a Mixture-of-Experts (MoE) \cite{zhou2022mixture} technique armed with a dynamic reinforcement learning-based routing strategy to maximize video violence recognition accuracy while minimizing computational costs incurred by using larger vision transformers. The evaluation results show that large pre-trained vision transformers are accurate models for video violence recognition. On top of that, experiments demonstrate that using the MoE technique and a cost-aware routing mechanism, the proposed model can achieve state-of-the-art results with a computational cost close to the most efficient vision transformers available.

\section{Proposed method}
\label{sec:proposed:method}

\begin{figure}[!htbp]
\centering
        \includegraphics[width=0.8\textwidth]{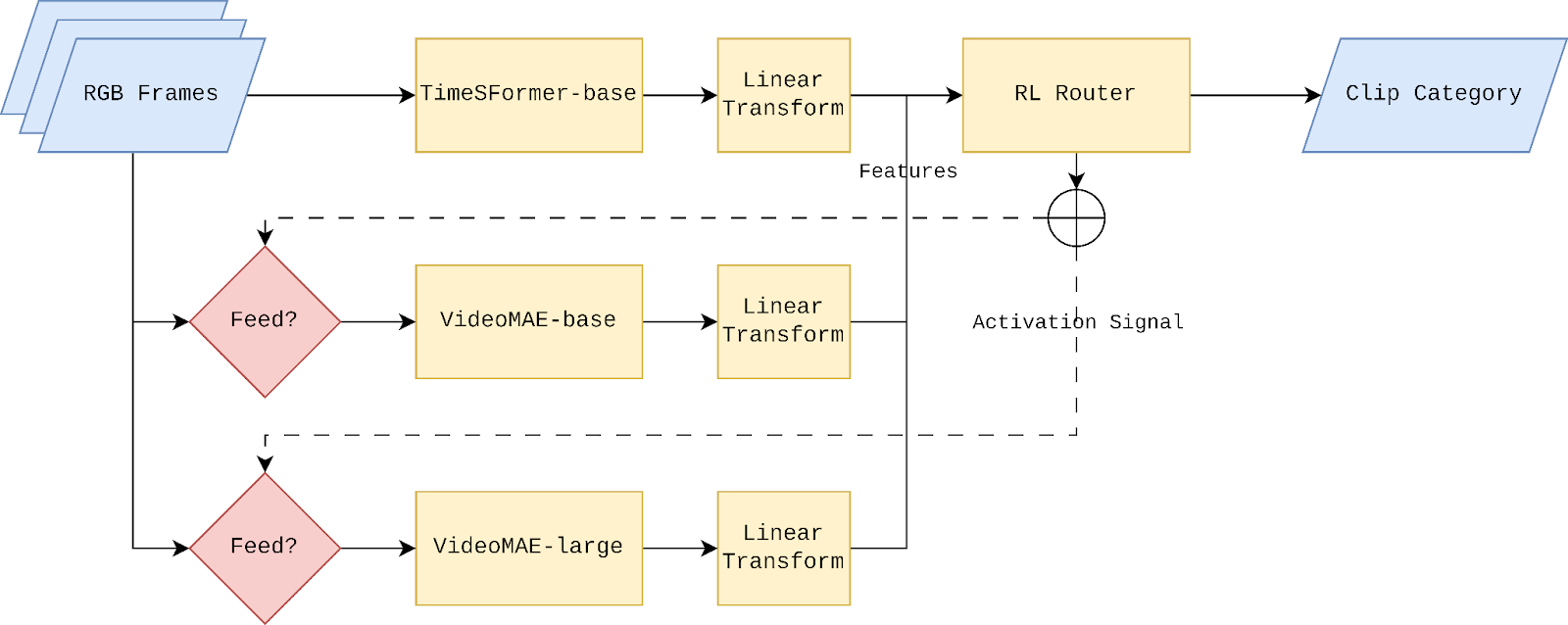}
    \caption{The architecture of the Sparse Recurrent Post-activated Mixture of Experts (SRPMoE) system for video violence recognition. At first, the raw RGB frames are encoded by one of the experts. On the basis of its learned policy and encoded observations of RGB frames, the RL router may activate additional experts.}
    \label{fig:SRPMoE}
\end{figure}

The proposed method consists of two main components: (i) Pre-trained vision transformers trained on the RWF dataset (i.e., experts). (ii) An MoE system using a novel reinforcement learning-based router and classifier model that activates each expert based on the current video clip and determines the clip category (Fight or Non-fight). Figure \ref{fig:SRPMoE} displays the general architecture of the proposed method. The reinforcement learning router observes the video clip (RGB frames) through expert embeddings. The RL router then decides whether to choose the clip category immediately or gather more insight by activating one of the larger vision transformers and using its richer embedding. The RL router finishes the classification process by choosing one of the possible categories for the input video clip (Fight or Non-Fight). The details of each component in Figure \ref{fig:SRPMoE} are discussed in the following sections.

The intuition behind an MoE architecture in this study is to achieve the highest accuracy possible while minimizing computational costs. As transformers are larger and more computationally demanding than previous models, the use of MoE architectures becomes more worthwhile. Glam \cite{du2022glam}, Switch Transformer \cite{fedus2022switch}, Multi-model MoE \cite{shen2023scaling}, Residual Mixture of Experts \cite{wu2022residual}, and RoME \cite{satar2022rome} are examples of such usecases. In our study, vision transformers’ accuracy and computational cost cover a wide range. While VideoMAE models achieve state-of-the-art accuracy on video classification tasks, they are significantly computationally heavier than TimeSFormer models. Nonetheless, TimeSFormer models can accurately classify videos in a large number of cases. For example, TimeSFormer-B can correctly predict video categories in the Kinetics 400 dataset 78\% of the time with 3.3\% of the computational cost of the leading VideoMAE model (TimeSFormer-B=0.6 TFLOPs vs. VideoMAE (ViT-H)=17.8 TFLOPs). Even VideoMAE with ViT-B backbone archives comparable accuracy on the Kinetics 400 (VideoMAE (ViT-B)=81.5\%, VideoMAE (ViT-L)=85.1\%) while having significantly less computational cost (VideoMAE (ViT-B)=8.9 TFLOPs, VideoMAE (ViT-H)=17.8 TFLOPs) \cite{bertasius2021space, tong2022videomae}. Therefore, larger models are only beneficial when smaller models are incapable of correctly classifying video clips.

\begin{figure}[!htbp]
\centering
        \includegraphics[width=1.0\textwidth]{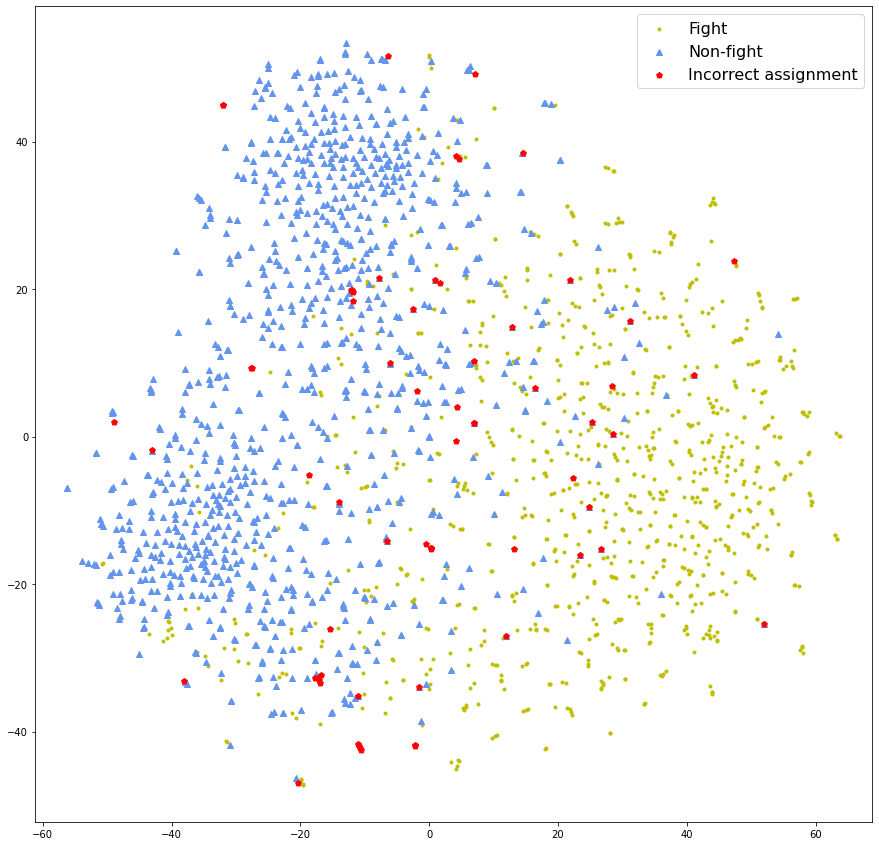}
    \caption{2D t-SNE visualization of the RWF dataset. Circles (green) represent fight videos and triangles (blue) represent non-fight videos. The points incorrectly classified as the opposite class are marked with pentagons (red). The visualization demonstrates that the majority of erroneous points are located at the intersection of the two categories.}
    \label{fig:error:visualization}
\end{figure}

Figure \ref{fig:error:visualization} shows the t-SNE \cite{van2008visualizing} representation of videos in the RWF dataset acquired using the TimeSFormer-B embeddings. The incorrectly classified videos by the TimeSFormer-B model in the evaluation subset of the RWF are displayed in red (pentagon). According to Figure \ref{fig:error:visualization} (i) These errors occur at the boundary between Fight and Non-fight clusters where the embeddings from the Fight and Non-fight classes cannot be clearly separated. (ii) The incorrect classifications occur on a small portion of videos. Combining the observations from Figure \ref{fig:error:visualization} and the points presented in the last paragraph, it is possible to conclude: (i) In order to reduce the model’s error in the boundary regions we can use higher accuracy models which are larger and have unsupervised pre-training. (ii) By delegating the classification task to a larger model in the error regions, the additional computational cost will be limited as the error region covers only a small portion of our representation space. In conclusion, the use of a cost-aware router in an MoE system based on vision transformer experts can result in an accurate video classification model while having an insignificant adverse effect on computational cost. The empirical evidence for this conclusion is presented in Section \ref{sec:experiments}.

MoE architecture can create a fair balance between accuracy and performance. As opposed to ensemble techniques which always substantially increase the computational cost of the learning system, MoE systems can be designed with limited computational overhead and reduced inference costs. Additionally, when compared with distillation methods which usually reduce the model's overall accuracy, MoE systems can be designed to enhance the system’s accuracy by incorporating knowledge from different models. The benefits of using an MoE system have naturally led to several applications in different fields of machine learning. As machine learning models continue to get larger, especially with the increased saturation cap of transformers, MoE architectures become more favorable. Deep models' large size encourages the use of MoE methods to reduce their computational costs while keeping their high accuracy.

The proposed MoE system in this study employs a reinforcement learning-based router. Expert activation for embedding extraction is performed by the RL-based router. This is possible using reinforcement learning and a post-activated router MoE system. Pre-activated router MoE is a common architecture in MoE systems. In such architectures, the router directly observes the input and only based on this observation, selects the set of experts to be activated for the current input. In contrast, in a post-activated router MoE, the router is placed after the experts and uses the experts’ output (i.e., embeddings) to decide the activation of experts. For example, the router in \cite{fedus2022switch} can select between different classification heads based on the generated embedding.

\begin{figure}[!htbp]
    \captionsetup[subfigure]{justification=Centering}

\begin{subfigure}[t]{1.0\textwidth}
    \includegraphics[width=\textwidth]{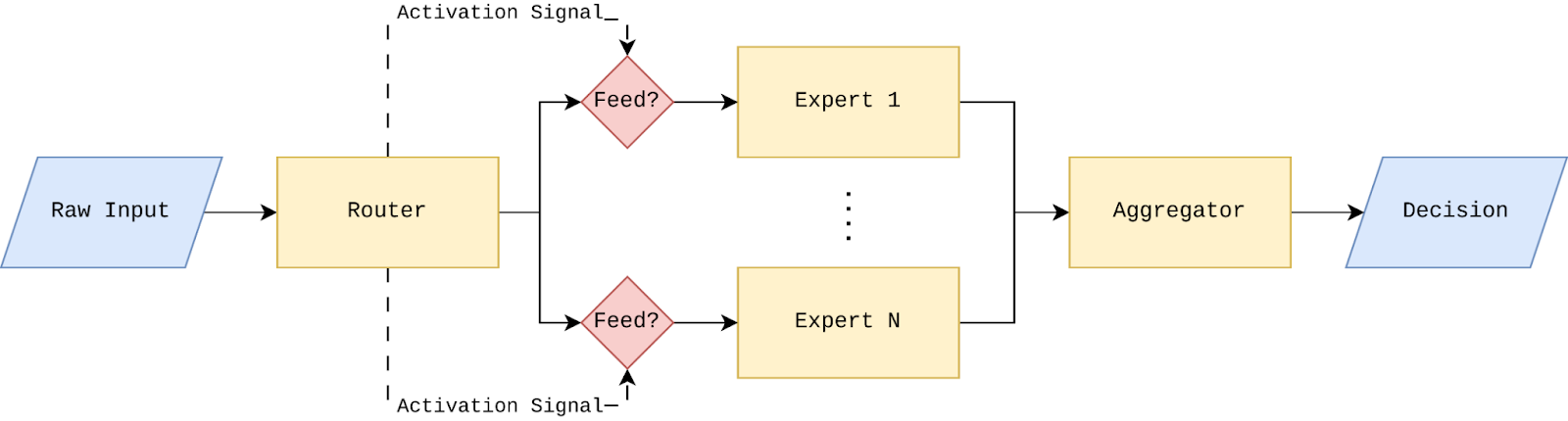}
    \caption{Design of a pre-activated Mixture of Experts. The router directly observes the raw input and selects experts solely based on raw observation. The final decision is made using a dedicated aggregator combining experts' embeddings or selections.}
    \label{fig:pre:moe}
\end{subfigure}
\bigskip 
\begin{subfigure}[t]{1.0\textwidth}
    \includegraphics[width=\linewidth]{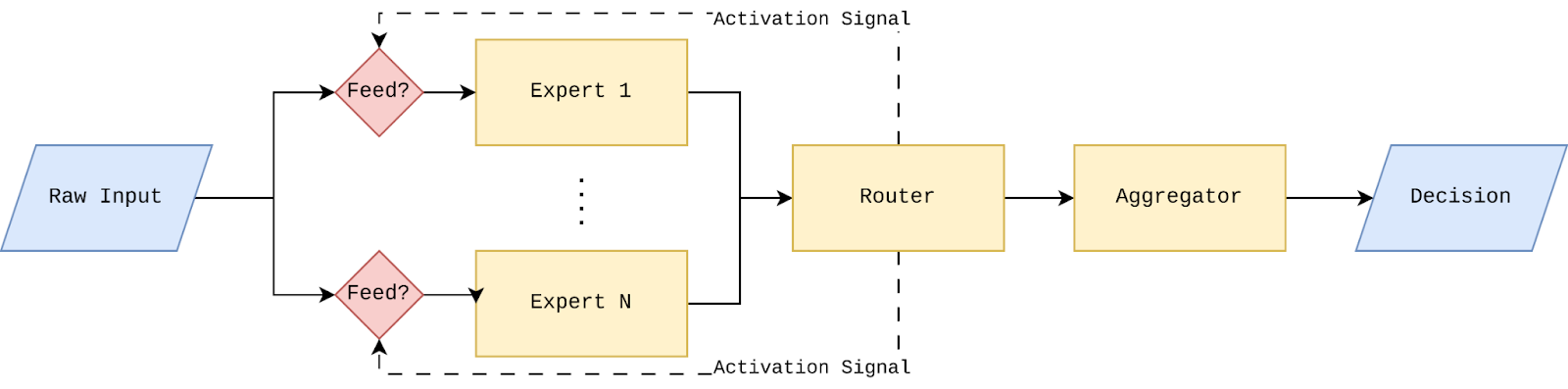}
    \caption{The general design of a post-activated mixture of experts system. The router observes the input indirectly through expert embeddings. The router or an aggregator makes the final decision.}
    \label{fig:post:moe}
\end{subfigure}

    \caption{Comparison between pre-activated (a) and post-activated (b) Mixture of Experts design.}
    \label{fig:pre:post:moe}
\end{figure}

The MoE system presented in this study (refer to Figure \ref{fig:SRPMoE}) is a special type of sparse post-activated MoE system we call Sparse Recurrent Post-activated MoE (SRPMoE). This technique of visual data processing is loosely inspired by the different visual pathways in the human brain \cite{van2021comparing}. The idea of a recurrent router is similar to the conditional routing of visual data in the brain to balance the latency and detail of extracted information \cite{pessoa2010emotion}. Therefore, the difference between vanilla MoE and recurrent MoE is the router’s ability to activate more experts based on feedback received from other experts. For example, the router might first select expert $E_{1}$ for embedding generation. Based on the generated embedding, it decides to engage more experts to gather more knowledge about the current input. The dynamic selection of experts in a feedback loop helps the router combine knowledge from different experts. It also stops expert activation when an accurate decision is possible using the currently gathered knowledge. Additionally, the activation cost of different experts can be considered in router training to minimize inference costs.

An SRPMoE architecture can be defined and solved using a reinforcement learning framework. To put the SRP router’s task into a reinforcement learning framework, the notion of classification validity, computational cost, and expert knowledge should be introduced to the standard RL formulation. The reward function is a critical part of every RL formulation. Equations \ref{eq:reward:total} to \ref{eq:reward:cost} show how rewards are formulated in this study.

\begin{equation}
\resizebox{.35\hsize}{!}{%
$R(o_{t},a_{t})=R_{c(o_{t})},a_{t})+\lambda R_{e}(a_{t})$%
}
\label{eq:reward:total}
\end{equation}

\begin{equation}
\resizebox{.4\hsize}{!}{%
$R_{c}(o_{t},a_{t}) = \begin{cases} +1 & \text{if } l_{a}(a_{t}) = l_{o}(o_{t}) \\ 0 & \text{if } l_{a}(a_{t}) = null \\ -1 & \text{if } l_{a}(a_{t}) \ne l_{o}(o_{t}) \end{cases}$%
}
\label{eq:reward:classification}
\end{equation}

\begin{equation}
\resizebox{.25\hsize}{!}{%
$R_e(a_t)=-1\times\frac{C(e_i)}{\sum_{j=1}^EC(e_j)}$%
}
\label{eq:reward:cost}
\end{equation}

Where $R(.)$ is the reward function, $R_{c}(.)$ is the classification reward, and $R_{e}(.)$ is the expert reward. The $o_t$ is the observation at time-step $t$, and $a_t$ is the action taken by the router at time-step $t$. The $l_{a}(.)$ function indicates the class label intended by the agent's action, which can be null if the action is to activate more experts. Additionally, the $l_{o}(.)$ function indicates the correct label for the observation. Finally, the $C(.)$ function is the computational cost of expert $e$. To determine the $R_e$, which depends on the router's action $a_t$, the computation cost of the selected expert, $C(e_i)$, is divided by the computation cost of all experts ($E$) for normalization.

Equation \ref{eq:reward:total} is the abstract definition of the reward function w.r.t. the current observation and action. This reward function is defined as the summation of the classification reward and the expert’s computational cost reward. The classification reward function produces a +1 reward when the label selected by the router’s action equals the true label of the observed video clip. In contrast, this function produces a -1 reward when the label selected by the router’s action does not equal the true label of the observed video clip. Moreover, when the router’s action is not a classification signal, the classification reward function returns a neutral (0) reward. The expert’s computational cost reward function always returns a negative reward, a regularized function of the expert's computational cost (GFLOPs). To control the degree of penalization of the router for its use of computing resources, the constant $\lambda$ is multiplied by the expert computational cost function. This is done when calculating the total reward value. The empirical analysis in section \ref{sec:experiments} shows the effects of this coefficient on SRPMoE accuracy and computational cost.

The described reward function produces reward signals for the router’s RL agent. These signals are later used in the Q-learning formulation shown in Equation \ref{eq:value:learning}. Expert embedding is introduced in the form of a function of the current observation (RGB frames). Based on the router's action, the embedding function generates the observation's embedding using different experts. The embedding function affects the advantage function or the action likelihood depending on the router’s RL algorithm.

\begin{equation}
\resizebox{.8\hsize}{!}{%
$\nabla_{\theta}J(\pi_{\theta})=E\left[\sum_{t=0}^T\nabla_{\theta}\log\pi_{\theta}(a_t|E_{a_t}(o_t))A^{\pi_{\theta}}(E_{a_t}(o_t),a_t)\right]$%
}
\label{eq:policy:learning}
\end{equation}

\begin{equation}
\begin{split}
    Q_t(E_{a_t}(o_t),a_t) &= Q_t(E_{a_t}(o_t),a_t) \\
    &+ \alpha\left[R(E_{a_t}(o_t),a_t) \right. \\
    &+ \left. \gamma\max Q_{t+1}(E_{a_{t+1}}(o_{t+1}),a_{t+1})-Q_t(E_{a_t}(o_t),a_t)\right]
\end{split}
\label{eq:value:learning}
\end{equation}

Where $Q_{t}(.)$ is the current estimation of Q value and $Q_{t+1}(.)$ is the estimated Q value for the next time-step. The model's observation is replaced by $E_{a_t}(o_t)$ which is the embedding that was generated by the selected expert. The $R(.)$ is the reward function defined in Equation \ref{eq:reward:total}. Moreover, $\alpha$ is the learning rate and $\gamma$ is the Q-learning discount factor.

The router is implemented using an extended version of the Deep Q Network (DQN) method called double dueling DQN with experience replay \cite{mnih2015human}. The choice of a value learning off-policy method for this task is because of empirical evidence that supports the superiority of the DQN model on computer vision tasks \cite{le2022deep}. Value-based RL methods have less inter-sample correlation and hence avoid sub-optimal solutions and overfitting in tasks with high-dimensional inputs \cite{le2022deep}. Additionally, value-based methods usually perform better on discrete action spaces than policy-gradient methods such as PPO \cite{le2022deep}. The proposed neural network-based DQN receives 768-dimensional input embeddings despite the difference in expert embedding size (i.e., VideoMAE ViT-L = 1024). This is possible using a learnable linear transformation mapping each expert’s embedding to the dimensions of the DQN agent's observation space.

The router is especially sensitive to expert overfitting as expert assignment is learned from the training dataset. As a result, if an expert performs artificially well on the training set it can create a false selection confidence in the router. To prevent such a scenario, in addition to minimizing the experts’ overfitting, regularization techniques are also applied to the router. Firstly, the input expert embeddings are augmented using Gaussian noise \cite{li2021simple} to create a larger set of training embeddings and reduce the deterministic nature of the router's observation. The effectiveness of these techniques besides standard regularization techniques like dropout is empirically demonstrated in the ablation analysis in section \ref{sec:experiments}.

\section{Experiments}
\label{sec:experiments}

The proposed SRPMoE method improves the current state-of-the-art vision transformers by introducing a cost-aware routing mechanism to minimize the computational cost of these models while maximizing the accuracy of the video violence classification system. A router's behavior is affected by the cost associated with each expert (vision transformer) in terms of risking computation for increased accuracy or reducing computational costs by only using the largest models when they are absolutely necessary. Additionally, the proposed SRPMoE has several moving parts suggested in this study which require careful evaluation.

RWF dataset \cite{cheng2021rwf} is the largest and most comprehensive publicly available video violence recognition dataset. The small size and the lack of variety of video clips in datasets such as Hockey Fights \cite{bermejo2011violence}, Movies \cite{gong2008detecting}, and Crowd Fights \cite{Hassneretal:SISM12} encouraged us to focus our experiments on the RWF dataset. This dataset contains 2000  surveillance video clips, collected from YouTube streaming service \footnote{https://www.youtube.com/}, equally divided into fight and non-fight (1000 fight and 1000 non-fight video clips). Additionally, the dataset has a pre-defined train and test split with 1600 clips in the training portion and 400 clips in the testing segments, which are equally divided between fight and non-fight clips. Video clips are 5 seconds long with 30 frames per second; hence, each video clip contains 150 unique frames. The video clips have various resolutions ranging from a minimum of 204x188 to a maximum of 1920x1080 with an average of 527x360. Each clip with a fight label depicts a physical fight between two or more people. The clips with a non-fight label contain normal (no visible fight) versions of the scenes in the fight clips.

The pre-trained vision transformers are fine-tuned on the RWF dataset with a linear classification head. Because the router uses the training dataset to learn expert assignment, keeping the model from overfitting on the training dataset is critical. To avoid overfitting, video augmentation techniques and early stopping are utilized. After fine-tuning, the linear classification heads are removed, so the fine-tuned models are only used as embedding functions. Table \ref{tab:benchmark:experts:rwf} presents the top 1 accuracy of the selected models on the RWF dataset.

\begin{table}[!htbp]
\centering
\caption{Accuracy, size, and computational cost of vision transformers on the RWF dataset. VideoMAE ViT-L achieved the highest test accuracy (92.4\%) and VideoMAE ViT-S achieved the lowest test accuracy (85.0\%).}
\label{tab:benchmark:experts:rwf}
\noindent
\begin{adjustbox}{width=\textwidth}
\begin{tabular}{llcc}
\toprule
\textbf{Class} &
\textbf{Model} &
\textbf{Test Accuracy (\%)} &
\textbf{\# Parameters (M)} \\
\midrule

\multirow{4}{2cm}{\textbf{Self-attention-based}}
& VideoMAE ViT-H & 92.3 & 633 \\
& VideoMAE ViT-L & 92.4 & 305 \\
& VideoMAE ViT-B & 91.8 & 67 \\
& VideoMAE ViT-S & 85.0 & 22 \\

& TimeSFormer-L & 88.0 & 121.3 \\
& TimeSFormer-HR & 89.3 & 121.7 \\
& TimeSFormer-B & 85.5 & 121.2 \\

\cmidrule(l){2-4}

\multirow{4}{2cm}{\textbf{CNN-based}}
& SSHA \cite{mohammadi2023video} & 90.4 & 13.7 \\
& SepConvLSTM \cite{islam2021efficient} & 89.7 & 0.3 \\
& SPIL \cite{su2020human} & 89.3 & N/A \\
& Flow Gated Network \cite{cheng2021rwf} & 87.2 & 0.2 \\

\bottomrule

\end{tabular}
\end{adjustbox}
\end{table}

VideoMAE ViT-L, VideoMAE-ViT-B, and TimeSFormer-B models are selected as expert models for the SRPMoE system. The selection criteria include models from a spectrum of accuracy and computational cost. Having models with different properties increases the diversity of choices available to the router. This improves its ability to optimize MoE system accuracy and performance. Additionally, as the models need to be loaded into the system’s memory, it is wise to keep the number of models available to the router minimal. Therefore VideoMAE ViT-L is at the accuracy end of the models’ spectrum with the highest accuracy on the RWF dataset (92.4\%). TimeSFormer-B is at the performance end of the model’s spectrum with the lowest computational cost (0.59 TFLOPs). Including the VideoMAE ViT-H is not valid as VideoMAE ViT-L achieves better accuracy (VideoMAE ViT-L=92.4\%, VideoMAE ViT-H=92.3\%) while having lower computational cost (VideoMAE ViT-L=8.9 TFLOPs, VideoMAE ViT-H=17.8 TFLOPs). The same rationale applies to selecting VideoMAE ViT-B over TimeSFormer-L and TimeSFormer-HR as it has higher accuracy with less computational cost.

The router is trained using frozen experts on the same train/test split as the experts. The router's MLP policy is based on a two-layer fully connected neural network with a hidden size of 64 neurons. The input embedding size of the router is 768, according to the embedding size of the experts after linear transformation. Additionally, the number of output nodes is determined by the number of experts. The hidden layer activation function of the router is Tanh to increase the stability of the model \cite{hwangbo2019learning}. As for output layer activation, the DQN router uses linear activation for value estimation. This is unlike the PPO router which uses softmax activation for action selection. The DQN router is trained for 50,000 episodes. Table \ref{tab:benchmark:SRPMoE:rwf} shows the accuracy and computation cost of the DQN-based SRPMoE system using different cost coefficients.

\begin{table}[!htbp]
\centering
\caption{Accuracy and computational cost of the SRPMoE model on the RWF dataset. An increase in the cost coefficient reduces the computational cost with the penalty of reduced accuracy.}
\label{tab:ablation}
\begin{adjustbox}{width=\textwidth}
\begin{tabular}{lcccc}
\toprule
\textbf{Model} &
\textbf{Cost Coefficient} &
\textbf{Train accuracy (\%)} &
\textbf{Test accuracy (\%)} &
\textbf{Avg. TFLOPs}\\
\midrule

\multirow{3}{2cm}{\textbf{SRPMoE (DQN)}}
& 0.0 & 94.3 & 92.2 & 3.38 \\
& 0.1 & 94.8 & 92.0 & 2.66 \\
& 0.2 & 94.5 & 91.9 & 2.36 \\
& 0.3 & 94.1 & 91.1 & 1.86 \\
& 0.4 & 93.2 & 90.7 & 1.53 \\
& 0.5 & 93.5 & 89.2 & 0.96 \\

\bottomrule
\end{tabular}
\end{adjustbox}
\end{table}

\begin{figure}[!htbp]
\centering
        \includegraphics[width=0.8\textwidth]{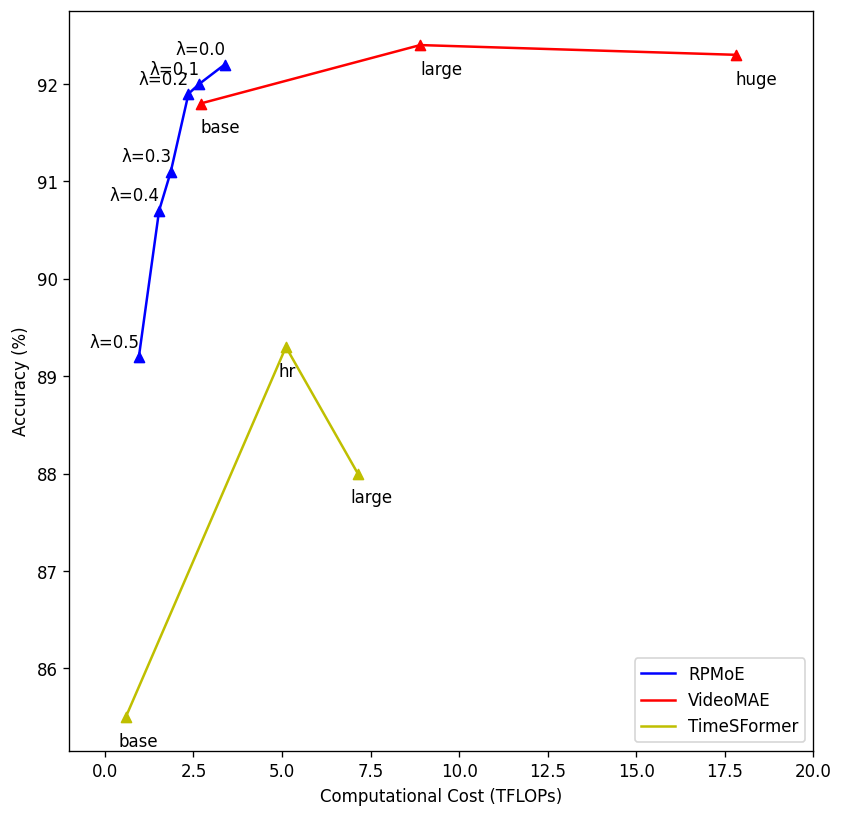}
    \caption{Comparison of accuracy and computational cost of SRPMoE, VideoMAE, and Timesformer models. The SRPMoE models are trained using different cost coefficients ranging from 0.0 to 0.5. Models closer to the axis's top-left are computationally more efficient by gaining more accuracy using less computational cost.}
    \label{fig:comparison}
\end{figure}

According to Table \ref{tab:benchmark:SRPMoE:rwf}, the SRPMoE system achieved an accuracy of 92.2\% with an average computational cost of 3.38 TFLOPs with a cost coefficient of 0. The SRPMoE in this configuration is close to the VideoMAE ViT-B (92.4\%) regarding its accuracy while having a smaller average computational cost (3.4 vs. 8.9 TFLOPs). The cost coefficient cost of 0 indicates that no negative rewards are received for using larger models in this setup. However, reinforcement learning models usually exploit shorter trajectories if the expected gains are similar, as shorter action sequences reduce reward uncertainty \cite{pacchiano2021dueling}. Therefore, the SRPMoE model with average computational cost reward uses the initial TimeSFormer-B representation when accurate classification is possible with high confidence. Increasing the cost coefficient discourages the use of large models, reducing the system’s average computational cost with a small accuracy decrement. For example, with a cost coefficient of 0.2, the SRPMoE system achieves an accuracy of 91.9\% with 2.36 TFLOPs of average computational cost. This accuracy is higher than the VideoMAE ViT-B model, with an accuracy of 91.8\% despite the smaller average computational cost (2.36 vs. 2.7 TFLOPs). Furthermore, using the cost coefficient of 0.5, the SRPMoE system achieves an accuracy of 89.2\% with an average computational cost of 0.96 TFLOPs. This is a significant gain over the TimeSFormer-B model’s accuracy (85.5\%). In contrast, the average computational cost difference between these models is relatively small compared to the large increase in computational cost considering the larger TimeSFormer and VideoMAE models. Figure \ref{fig:comparison} clearly shows the superior accuracy vs. performance trade-off management of the SRPMoE system. Points closer to the top left of the accuracy/computation cost space have higher accuracy per unit of computational cost.

\begin{figure}[!htbp]
    \captionsetup[subfigure]{justification=Centering}

\begin{subfigure}[t]{0.5\textwidth}
    \includegraphics[width=\textwidth]{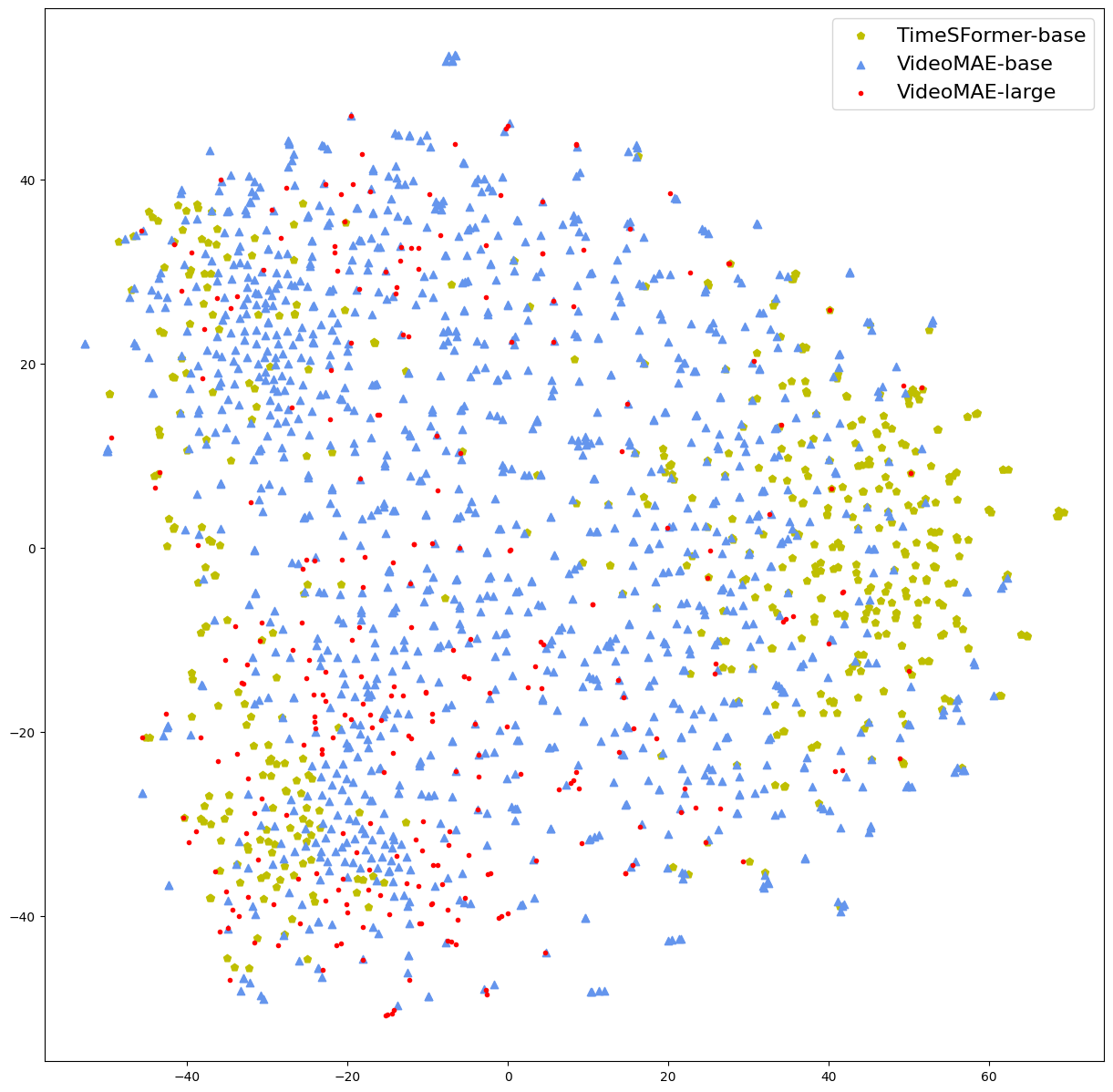}
    \caption{$\lambda = 0.0$.}
    \label{fig:cc:0}
\end{subfigure}
\begin{subfigure}[t]{0.5\textwidth}
    \includegraphics[width=\linewidth]{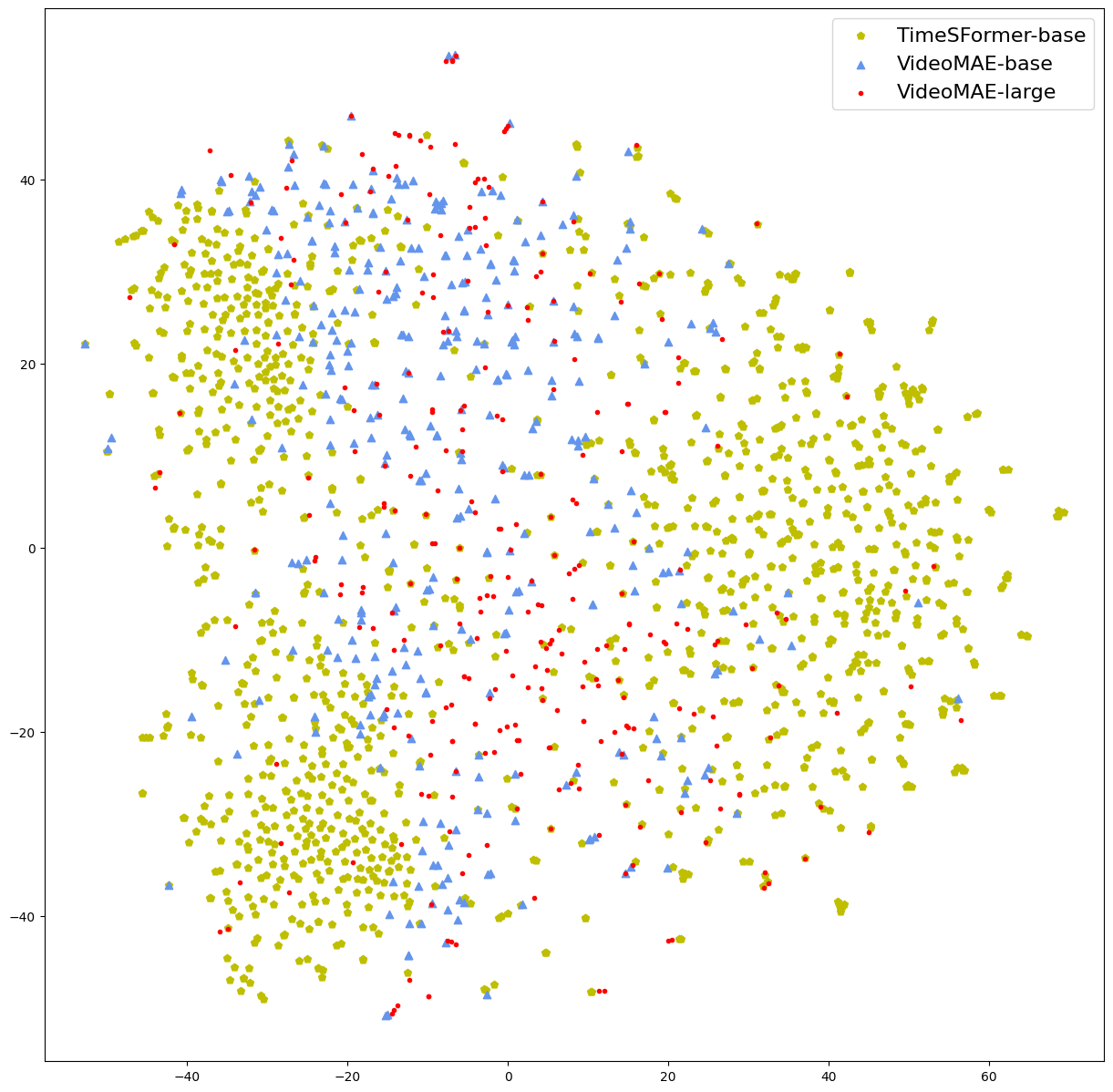}
    \caption{$\lambda = 0.2$.}
    \label{fig:cc:2}
\end{subfigure}
\bigskip 
\begin{subfigure}[t]{0.5\textwidth}
    \includegraphics[width=\linewidth]{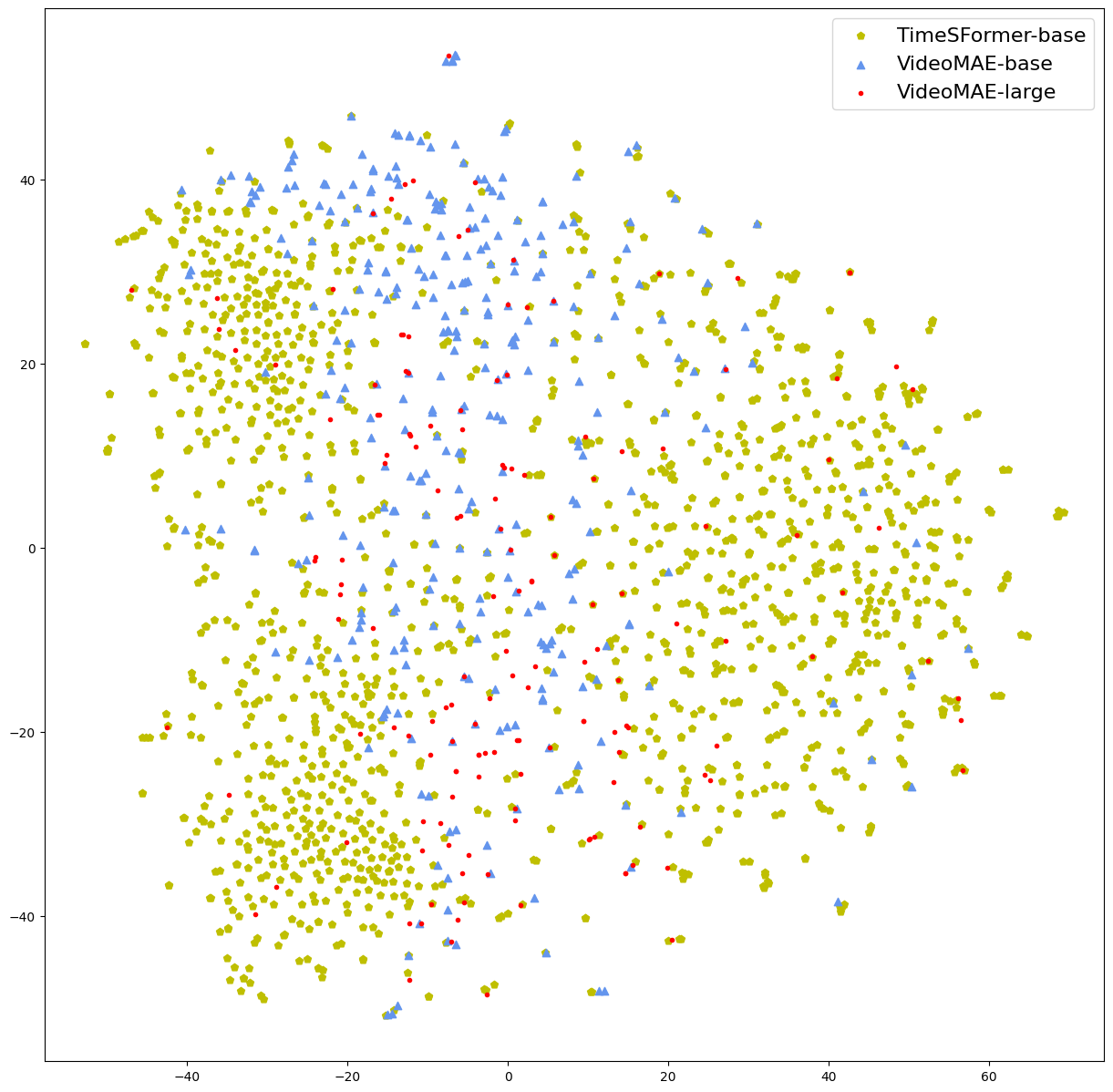}
    \caption{$\lambda = 0.4$.}
    \label{fig:cc:4}
\end{subfigure}
\begin{subfigure}[t]{0.5\textwidth}
    \includegraphics[width=\linewidth]{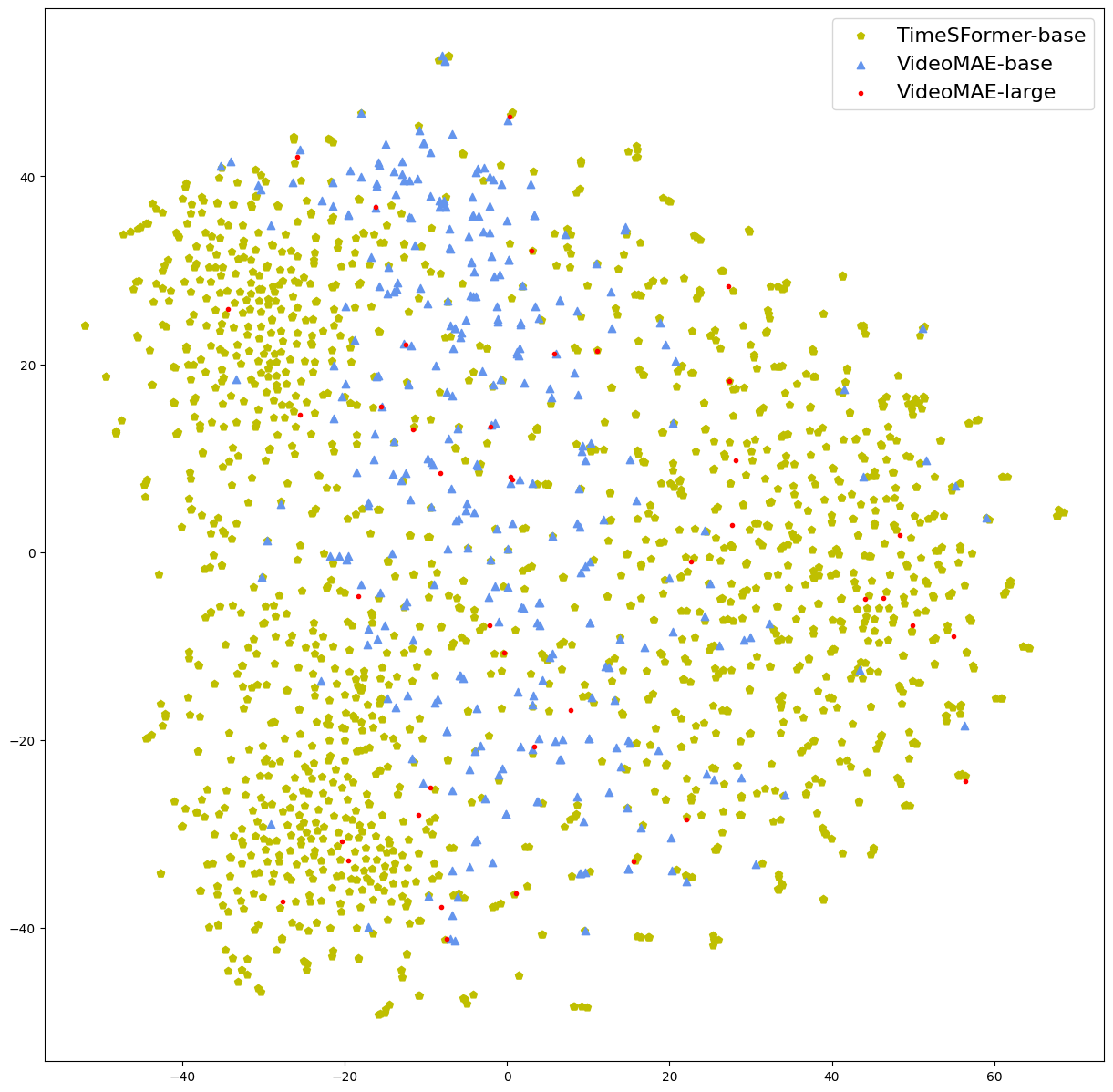}
    \caption{$\lambda = 0.5$.}
    \label{fig:cc:5}
\end{subfigure}

    \caption{2D t-SNE visualization of experts selected by the SRPMoE router for each RWF dataset video. Pentagons (green) represent the points assigned to the TimeSFormer-base model, triangles (blue) are assigned to the VideoMAE-base model, and circles (red) are assigned to the VideoMAE-large model. Gradual increase in cost coefficient ($\lambda$) from figure \ref{fig:cc:0} to \ref{fig:cc:5} shows the gradual reduction of large vision transformer models. Additionally, the SRPMoE router prioritizes the use of large vision transformers in the high-uncertainty regions of the intersection between classes.}
    \label{fig:cc}
\end{figure}

The cost coefficient can be interpreted as the allowed computational risk. Most incorrect classifications of a model occur at the edge of the class clusters. As a result, using larger and more accurate models in such regions reduces classification correctness uncertainty. However, increasing intolerance toward computational costs makes smaller models in higher-risk regions profitable. According to Figures \ref{fig:cc:0} to \ref{fig:cc:5}, gradually increasing the cost coefficient forces the router to rely more on smaller models in high-risk areas between the two clusters.

\subsection{Ablation analysis}

Design choices of the SRPMoE can drastically change the model’s behavior as reinforcement learning models are more sensitive to model architecture and hyperparameter selection than supervised methods. The empirical evaluation of the main design choices shows the validity and difference of the possible varieties of the SRPMoE system. The train and test accuracy and average computational cost of variations of the SRPMoE designs with regards to different cost coefficients are presented in Table \ref{tab:ablation}.

\subsubsection{DQN vs. PPO}

The use of the PPO reinforcement learning algorithm results in a sub-optimal solution for balancing the router's accuracy and computational cost. According to Table \ref{tab:benchmark:SRPMoE:rwf} and \ref{tab:ablation}, the SRPMoE PPO algorithm, on average, achieves 1.6\% lower train accuracy and 3.9\% lower test accuracy compared to its DQN counterpart. This result validates the fact that PPO has a higher inter-sample correlation and hence overfits the training set. Additionally, the PPO algorithm is more sensitive to the cost coefficient because after adding the negative computational cost reward, the system's tolerance toward computational cost drastically increases, resulting in a loss of accuracy. This is interesting as the computational cost when the cost coefficient is zero is similar to the SRPMoE DQN.

\subsubsection{Feature aggregation}

Another design choice is whether to use direct embedding or aggregated embedding when using more than one expert for video classification. In principle, feature aggregation can result in the combined information of multiple experts. However, this is not always true in practice as combining highly correlated and overlapping features can result in model bias and overfitting \cite{mungoli2023adaptive}. This is apparent in Table \ref{tab:ablation} data as the SRPMoE system that rely on aggregated features achieves 0.81\% and 1.28\% less accuracy on the training and testing RWF subsets, respectively, compared to the direct embedding system.

\subsubsection{Expert embedding augmentation}

Expert embedding augmentation is effective when analyzing the presence and absence of this technique. In addition to increasing the router’s observation dataset size, removing embedding augmentation from the router’s training process results in over-reliance on the router on larger models. While training the SRPMoE without embedding augmentation results in an average increase of 1.8\% in train accuracy, the router’s test accuracy suffers a 2.38\% loss on average. The empirical results show that expert embedding augmentation can effectively reduce reinforcement learning-based router overfitting on the training subset.

\subsubsection{Expert overfitting}

Overfitted expert avoidance is critical in the SRPMoE design as region selection based on overfitted expert embeddings can distort the router’s understanding of each expert’s capability. This results in overconfidence in models with artificially high training performance, especially for larger models. Therefore, regularization of expert training loss and accuracy regardless of its evaluation performance is essential. Even when the evaluation accuracy is comparable to the most accurate variants of the expert, a large gap between the train and test accuracy can critically impair the router's region selection ability. In the overfit expert variant of the SRPMoE system, the early stopping technique is not utilized. Thus, the train accuracy of experts is near 100\% while evaluation accuracy is lower. This variant's train accuracy is on average 3.53\% higher than the SRPMoE system with non-overfit experts despite suffering only a 0.33\% loss of average test accuracy. However, the significant effect of using overfit experts is the artificial superiority of the larger models when trained on a dataset of limited size. This is why routers tend to favor larger experts when experts are overfitted. This is empirically demonstrated by the 3.7 TFLOPs increase in the average computational cost of the SRPMoE system when using overfit experts.

\begin{table}[!htbp]
\centering
\caption{Accuracy and computational cost of various SRPMoE designs.}
\label{tab:benchmark:SRPMoE:rwf}
\begin{adjustbox}{width=\textwidth}
\begin{tabular}{lcccc}
\toprule
\textbf{Model} &
\textbf{Cost Coefficient} &
\textbf{Train accuracy (\%)} &
\textbf{Test accuracy (\%)} &
\textbf{Avg. TFLOPs}\\
\midrule

\multirow{3}{4cm}{\textbf{SRPMoE (PPO)}}
& 0.0 & 94.8 & 90.2 & 3.47 \\
& 0.1 & 93.8 & 88.2 & 2.03 \\
& 0.2 & 92.9 & 86.7 & 1.1 \\
& 0.3 & 91.3 & 86.5 & 0.94 \\
& 0.4 & 91.6 & 86.2 & 0.84 \\
& 0.5 & 90.0 & 85.7 & 0.76 \\

\cmidrule(l){1-5}

\multirow{3}{4cm}{\textbf{SRPMoE (aggregation)}}
& 0.0 & 93.7 & 91.4 & 3.06 \\
& 0.1 & 94.6 & 90.4 & 2.61 \\
& 0.2 & 94.4 & 90.3 & 2.44 \\
& 0.3 & 93.3 & 89.7 & 1.05 \\
& 0.4 & 91.3 & 89.6 & 1.67 \\
& 0.5 & 92.2 & 88.0 & 1.11 \\

\cmidrule(l){1-5}

\multirow{4}{4cm}{\textbf{SRPMoE (w.o. feature augmentation)}}
& 0.0 & 95.1 & 89.6 & 5.31 \\
& 0.1 & 95.5 & 89.5 & 3.62 \\
& 0.2 & 96.1 & 89.2 & 2.31 \\
& 0.3 & 96.3 & 88.6 & 1.61 \\
& 0.4 & 96.2 & 88.2 & 1.49 \\
& 0.5 & 96.5 & 87.7 & 0.92 \\

\cmidrule(l){1-5}

\multirow{3}{4cm}{\textbf{SRPMoE (overfitted expert)}}
& 0.0 & 98.3 & 91.9 & 6.39 \\
& 0.1 & 98.2 & 91.6 & 6.31 \\
& 0.2 & 97.4 & 91.0 & 5.98 \\
& 0.3 & 97.5 & 90.9 & 5.93 \\
& 0.4 & 97.1 & 90.0 & 5.39 \\
& 0.5 & 97.1 & 89.7 & 4.99 \\

\bottomrule
\end{tabular}
\end{adjustbox}
\end{table}


\section{Discussion}
\label{sec:discussion}

Video transformers provide promising solutions to video violence recognition. Regarding the accuracy of video violence recognition models, vision transformers are superior to CNN-based models mainly through their higher learning capacity when trained, supervised, and self-supervised on larger datasets. Furthermore, although larger video transformers demand more computational cost, they present higher learning capacity per unit of computational cost, considering their learning capacity. For example, SlowFast requires 7 TFLOPs of computational resources with 34.3 million parameters, while TimeSFormer-L requires 7.1 TFLOPs of computational resources with 121.3 million parameters. This is especially useful as large-capacity models lead machine learning benchmarks, and keeping computational costs manageable is becoming a top priority.

The experiments demonstrate the viability of the SRPMoE system for a low overhead balance between accuracy and computational cost. The balance between accuracy and computational cost can be quantified and compared by introducing an average accuracy (percentage) over computational cost (TFLOPs) measure. The SRPMoE archives an average accuracy over computational cost ratio of 42.9 percent/TFLOPs. In comparison, TimeSFormer and VideoMAE  models achieve an average accuracy over computational cost of 20.4 and 9.4 percent/TFLOPs, respectively. Moreover, SRPMoE could be tuned to achieve higher or lower accuracy over computational cost values by tweaking the cost coefficient in the router’s reward formulation. This value can range from 27.2 for the cost coefficient of 0 to 92.9 for the cost coefficient of 0.5. In conclusion, SRPMoE provides a full range of solutions for video violence recognition with flexible prioritization between accuracy and computational cost.

A limiting factor of SRPMoE systems is the number of active experts available to the router. Although modern deep learning acceleration hardware can easily fit multiple models into their memories, the SRPMoE system memory management requires special attention. Memory management problems can be effectively managed by an effective selection of experts and using common memory management techniques such as lazy weight loading. Moreover, reinforcement learning models are more sensitive to model architecture, hyperparameter selection, and overfitting. Therefore, a careful search for possible design choices is essential to achieve optimal solutions.

\section{Conclusions}
\label{sec:conclusions}

This study introduces a novel Sparse Recurrent Post-activated Mixture of Experts (SRPMoE) violence recognition system. The proposed SRPMoE method uses a semi-supervised reinforcement learning-based routing mechanism to use large and computationally heavy video transformers in combination with efficient video transformers, reducing computational costs while maximizing violence recognition accuracy. The RL-based router is lightweight and requires no annotations on top of video-level labels. Moreover, using a cost-aware router, the proposed method allows for a dynamic balance between accuracy and computational cost. This makes this method suitable for a wide range of applications with different accuracy and computational cost requirements.

The use of layer-level routing can further decrease the computational cost of the proposed SRPMoE by skipping layers or making faster routing decisions based on low-level features. Additionally, previous studies have shown the importance of frame sampling in the accuracy of video action recognition models \cite{zheng2020dynamic}. Using a frame-level router for frame and expert selection is another interesting continuation of this study. Furthermore, the use of recent regularization methods in reinforcement learning \cite{wang2020improving} can increase router stability and test accuracy.

\section*{Acknowledgement}
\label{sec:acknowledgement}

We would like to express our gratitude toward the HuggingFace team\footnote{https://huggingface.co/} especially the transformers library \cite{wolf-etal-2020-transformers} for their open-source implementation of vision transformers. We also want to thank the stable-baselines\footnote{https://github.com/DLR-RM/stable-baselines3} library team for their implementation of common reinforcement learning algorithms \cite{stable-baselines3} which is used for rapid prototyping of reinforcement learning agents in this study.

\bibliographystyle{elsarticle-num} 
\bibliography{cas-refs}

\end{document}